\documentclass[11pt,a4paper]{article}
\usepackage{times,latexsym}
\usepackage{url}
\usepackage[T1]{fontenc}

\usepackage[ruled,vlined]{algorithm2e}
\usepackage{subfigure}
\usepackage{listings}
\usepackage{graphicx}
\usepackage{mathtools}
\usepackage{tabularx}
\usepackage{amsmath}
\usepackage{amssymb}
\usepackage{booktabs}
\usepackage{arydshln}
\usepackage{hyperref}
\usepackage[boldmath,np]{numprint}
\usepackage{cleveref}

\crefformat{section}{\S#2#1#3}
\crefformat{subsection}{\S#2#1#3}
\crefformat{subsubsection}{\S#2#1#3}
\crefrangeformat{section}{\S\S#3#1#4 to~#5#2#6}
\crefmultiformat{section}{\S\S#2#1#3}{ and~#2#1#3}{, #2#1#3}{ and~#2#1#3}

\newcommand{\npbold}{\fontseries{b}\selectfont}

\usepackage[acceptedWithA]{tacl2021v1}
\usepackage{tacl2021v1}

\usepackage{xspace,mfirstuc,tabulary}

\newif\iftaclinstructions
\taclinstructionsfalse 
\iftaclinstructions
\renewcommand{\confidential}{}
\renewcommand{\anonsubtext}{(No author info supplied here, for consistency with
\newcommand{\instr}
\fi

\iftaclpubformat 
\newcommand{\taclpaper}{final version\xspace}
\newcommand{\taclpapers}{final versions\xspace}
\newcommand{\Taclpaper}{Final version\xspace}
\newcommand{\Taclpapers}{Final versions\xspace}
\newcommand{\TaclPapers}{Final Versions\xspace}
\else
\newcommand{\taclpaper}{submission\xspace}
\newcommand{\taclpapers}{{\taclpaper}s\xspace}
\newcommand{\Taclpaper}{Submission\xspace}
\newcommand{\Taclpapers}{{\Taclpaper}s\xspace}
\newcommand{\TaclPapers}{Submissions\xspace}
\fi

\newcommand{\x}{\mathbf{x}}
\newcommand{\e}{\mathbf{e}}
\newcommand{\aaa}{\mathbf{a}}
\newcommand{\cc}{\mathbf{c}}
\newcommand{\w}{\mathbf{w}}
\newcommand{\bell}{{\boldsymbol\ell}}
\newcommand{\vv}{\mathbf{v}}
\newcommand{\h}{\mathbf{h}}
\newcommand{\y}{\mathbf{y}}
\newcommand{\rr}{\mathbf{r}}
\newcommand{\z}{\mathbf{z}}
\newcommand{\f}{\mathbf{f}}
\newcommand{\X}{\mathbf{X}}
\newcommand{\I}{\mathbf{I}}
\newcommand{\V}{\mathbf{V}}
\newcommand{\K}{\mathbf{K}}
\newcommand{\B}{\mathbf{B}}
\newcommand{\D}{\mathbf{D}}
\newcommand{\E}{\mathbf{E}}
\newcommand{\PP}{\mathbf{P}}
\newcommand{\W}{\mathbf{W}}
\newcommand{\HH}{\mathbf{H}}
\newcommand{\Sim}{\mathbf{S}}
\newcommand{\LL}{\mathbf{L}}
\newcommand{\UU}{\mathbf{U}}
\newcommand{\kk}{\mathbf{k}}
\newcommand{\bb}{\mathbf{b}}
\newcommand{\ii}{\mathbf{i}}
\newcommand{\jj}{\mathbf{j}}
\newcommand{\btheta}{\boldsymbol\theta}
\newcommand{\bphi}{\boldsymbol\phi}
\newcommand{\bPhi}{\boldsymbol\Phi}
\newcommand{\grad}[2]{\dfrac{\partial #1}{\partial #2}}
\DeclareMathOperator*{\argmax}{argmax}
\DeclareMathOperator*{\argmin}{argmin}

\title{Modelling Emotion Dynamics in Song Lyrics with State Space Models}

\author{
  Yingjin Song$^\diamond$\Thanks{This work was completed when the first author was at The University of Melbourne.} 
  \and
  Daniel Beck$^\dagger$
  \\
  \ \\
  $^\diamond$Department of Information and Computing Sciences, Utrecht University
   \\
  $^\dagger$School of Computing and Information Systems, The University of Melbourne
  \\
  \texttt{y.song5@uu.nl \space \space d.beck@unimelb.edu.au}
}

\date{}

\begin{document}
\maketitle
\begin{abstract}
  Most previous work in music emotion recognition assumes a single or a few song-level labels for the whole song. While it is known that different emotions can vary in intensity within a song, annotated data for this setup is scarce and difficult to obtain. In this work, we propose a method to predict emotion dynamics in song lyrics {\em without song-level supervision}. We frame each song as a time series and employ a State Space Model (SSM), combining a sentence-level emotion predictor with an Expectation-Maximization (EM) procedure to generate the full emotion dynamics. Our experiments show that applying our method consistently improves the performance of sentence-level baselines without requiring any annotated songs, making it ideal for limited training data scenarios. Further analysis through case studies shows the benefits of our method while also indicating the limitations and pointing to future directions.


  

\end{abstract}

\section{Introduction}
\label{sec:intro}
Music and emotions are intimately connected, with almost all music pieces being created to express and induce emotions \citep{juslin2004expression}. 
As a key factor of how music conveys emotion, lyrics contain part of the semantic information that the melodies cannot express \citep{besson1998singing}.
Lyrics-based music emotion recognition has attracted increasing attention driven by the demand to process massive collections of music tracks automatically, which is an important task to streaming and media service providers \citep{kim2010music, malheiro2016emotionally, agrawal2021transformer}.

\begin{figure}[t!]
  \centering
\includegraphics[width=\linewidth]{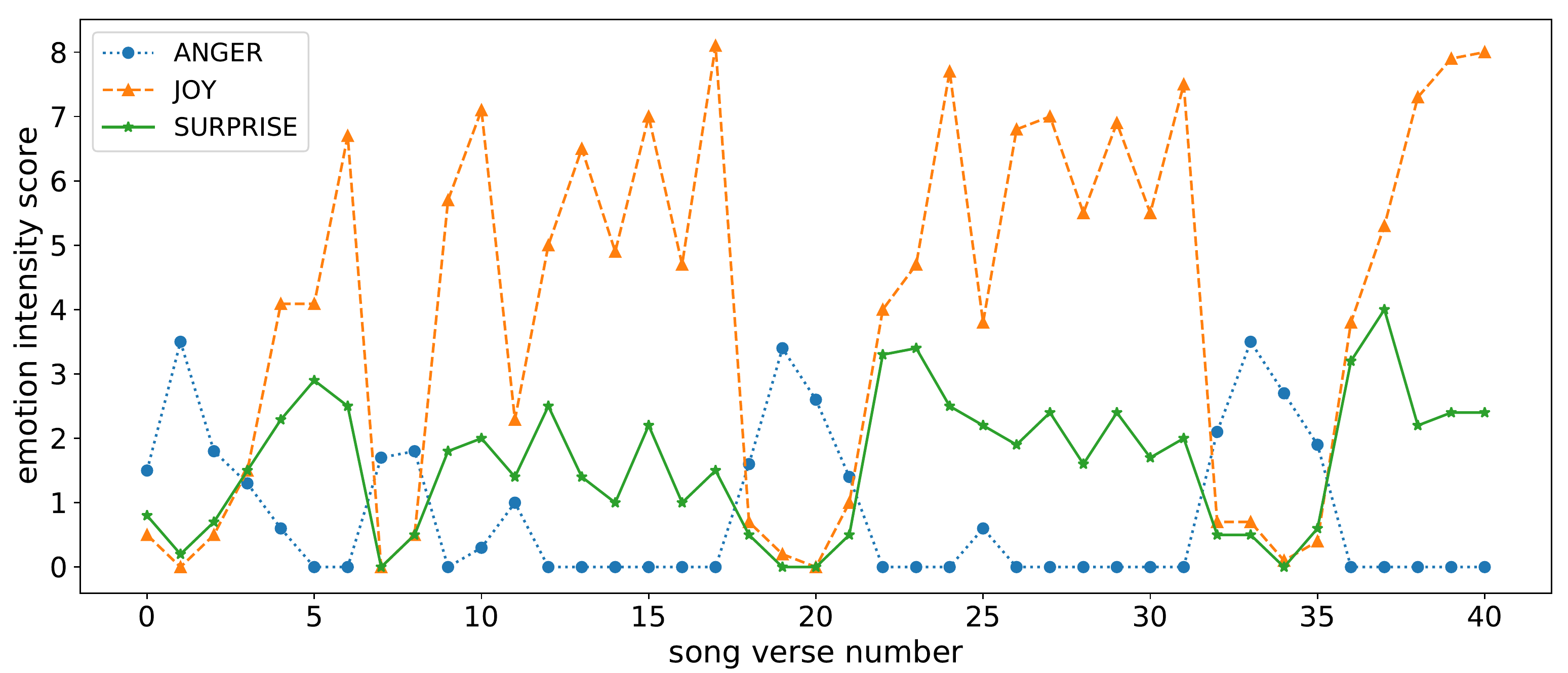}
  \caption{An illustration of emotion dynamics of a song in the {\sc LyricsEmotions} dataset of \newcite{mihalcea2012lyrics}. Note the intensities of each emotion vary from verse to verse within the song.}
  \label{fig:lyricexample}
\end{figure}

Most emotion recognition studies in Natural Language Processing (NLP) assume the text instance expresses a static and single emotion \citep{mohammad2017wassa, nozza2017multi, mohammad2018semeval}.
However, emotion is non-static and highly correlated with the contextual information, which makes the single-label assumption too simplistic in dynamic scenarios, not just in music \citep{schmidt2011modeling} but also in other domains such as conversations \citep{poria2019emotion}. Figure \ref{fig:lyricexample} shows an example of this dynamic behaviour, where the intensities of three different emotions vary within a song. Accurate emotion recognition systems should ideally be able to generate the full emotional dynamics for each song, as opposed to simply predicting a single label.
A range of datasets and corpora for modelling dynamic emotion transitions has been developed in the literature 
\citep{mckeown2011semaine, li-etal-2017-dailydialog, hsu-etal-2018-emotionlines, poria-etal-2019-meld, firdaus-etal-2020-meisd}, but most of them do not use song lyrics as the domain and assume discrete, categorical labels for emotions (either the presence or absence of one emotion).
To the best of our knowledge, the dataset from \newcite{mihalcea2012lyrics} is the only one that provides full fine-grained emotion intensity annotations for song lyrics at the verse\footnote{According to \citet{mihalcea2012lyrics}, a ``verse'' is defined as a sentence or a line of lyrics. } level. The lack of large-scale datasets for this task poses a challenge for traditional supervised methods. While previous work proposed methods for the similar sequence-based emotion recognition task, they all assume the availability of some levels of annotated data at training time, from full emotion dynamics \citep{DBLP:conf/aaai/KimLLP15} to coarse, discrete document-level labels \citep{tackstrom-mcdonald-2011-semi}.

The data scarcity problem motivates our main research question: {\em ``Can we predict emotion dynamics in song lyrics without requiring annotated lyrics?''}. In this work, we claim that the answer is affirmative. To show this, we propose a method consisting of two major stages: (1) a sentence or {\em verse-level} regressor that leverages existing emotion lexicons, pre-trained language models and other sentence-level datasets and (2) a State Space Model (SSM) 
that 
constructs a full {\em song-level} emotional dynamics given the initial verse-level scores. 
Intuitively, we treat each verse as a time step and the emotional intensity sequence as a latent time series that is inferred without any song-level supervision, directly addressing the limited data problem. To the best of our knowledge, {\em this scenario was never addressed before either for song lyrics or other domains}.



To summarize, our main contributions are:
\begin{itemize}
    \item We propose a hybrid approach for verse-level emotion intensity prediction that combines emotion lexicons with a pre-trained language model (BERT \citep{devlin-2019-bert} used in this work), which is trained on available sentence-level data.
    \item We further show that by using SSMs to model the song-level emotion dynamics, we can improve the performance of the verse-level approach without requiring any annotated lyrics. 
    \item We perform a qualitative analysis of our best models, highlighting its limitations and pointing to directions for future work.
\end{itemize}

\section{Background and Related Work}


\paragraph{Emotion Models.}

Human emotion is a long-standing research field in psychology, with many studies aiming at defining a taxonomy for emotions. 
In NLP, emotion analysis mainly employs the datasets which are annotated based on the {\em categorical} or the {\em dimensional} model.

The categorical model assumes a fixed set of discrete emotions which can vary in intensity. 
Emotions can overlap but are assumed to be separate ``entities'' for each other, such as {\em anger}, {\em joy} and {\em surprise}. 
Taxonomies using the categorical model include Ekman’s basic emotions \citep{ekman1993facial} and Plutchik's wheel of emotions \citep{plutchik1980general}. 
The dimensional models place emotions in a continuous space: the VAD (\textit{Valence}, \textit{Arousal} and \textit{Dominance}) taxonomy of \citet{russell1980circumplex} is the most commonly used in NLP. In this work, we focus on the Ekman taxonomy for purely experimental purposes, as it is the one used in the available data we employ. However, our approach is general and could be applied to other taxonomies.

\paragraph{Dynamic Emotion Analysis.}

Emotion Recognition in Conversation (ERC) \citep{poria2019emotion}, which focuses on tracking dynamic shifts of emotions, is the most similar task to our work.
Within a conversation, the emotional state of each utterance is influenced by the previous state of the party and the stimulation from other parties \citep{li-etal-2020-hitrans, ghosal2021exploring}.
Such an assumption of the real-time dynamic emotional changes also exists in music: the affective state of the current lyrics verse is correlated with the state of the previous verse(s) as a song progresses.

Contextual information in the ERC task is generally captured by deep learning models, which can be roughly categorized into sequence-based and graph-based methods \citep{hu-etal-2021-dialoguecrn}.
Sequence-based methods encode conversational context features using established methods like Recurrent Neural Networks \citep{poria-etal-2017-context, hazarika-etal-2018-icon, hazarika-etal-2018-conversational, majumder2019dialoguernn, hu-etal-2021-dialoguecrn} and Transformer-based architectures \citep{zhong-etal-2019-knowledge, li-etal-2020-hitrans}. They also include more advanced and tailored methods like Hierarchical Memory Network \citep{jiao2020real}, 
Emotion Interaction Network \citep{lu-etal-2020-iterative} 
and Causal Aware Network \citep{DBLP:conf/ijcai/ZhaoZL22}.
Graph-based methods apply specific graphical structures to model dependencies in conversations \citep{ghosal2019dialoguegcn, zhang2019modeling, lian2020conversational, ishiwatari-etal-2020-relation, shen-etal-2021-directed} by utilizing Graph Neural Networks \citep{kipf2017semisupervised}. 
In contrast to these methods, we capture contextual information using a State Space Model, mainly motivated by the need for a method that can train without supervision. Extending and/or combining an SSM with a deep learning model is theoretically possible but non-trivial, and care must be taken in a low-data situation such as ours.



The time-varying nature of music emotions has been investigated in music information retrieval \citep{caetano2012role}. To link the human emotions with the music acoustic signal, the emotion distributions were modelled as 2D Gaussian distributions in the Arousal-Valence (A-V) space, which were used to predict A-V responses through multi-label regression (\citeauthor{schmidt2010feature}, \citeyear{schmidt2010feature}; \citeauthor{schmidt2010prediction}, \citeyear{schmidt2010prediction}).  
Building on previous studies, \citet{schmidt2011modeling} applied structured prediction methods to model complex emotion-space distributions as an A-V heatmap. These studies focus on the mapping between emotions and acoustic features/signals, while our work focuses on the lyrics component. \citet{wu2014music} developed a hierarchical Bayesian model that utilized both acoustic and textual features, but it was only applied to predict emotions as discrete labels (presence or absence) instead of fine-grained emotion intensities as in our work.


\paragraph{Combining pre-trained Language Models with External Knowledge.}
Pre-trained language models (LMs) including BERT \citep{devlin-2019-bert}, XLNet \citep{yang2019xlnet} and GPT \citep{tom2020gpt} have achieved state-of-the-art performance in numerous NLP tasks. 
Considerable effort has been made towards combining context-sensitive features of LMs with factual or commonsense knowledge from structured sources, including domain-specific knowledge \citep{ying2019improving}, structured semantic information \citep{zhang2020semantics}, language-specific knowledge (\citeauthor{alghanmi2020combining}, \citeyear{alghanmi2020combining}; \citeauthor{de2021emotional}, \citeyear{de2021emotional}) and linguistic features (\citeauthor{koufakou2020hurtbert}, \citeyear{koufakou2020hurtbert}; \citeauthor{mehta2020bottom}, \citeyear{mehta2020bottom}).
This auxiliary knowledge is usually infused into the architecture by concatenating them with the Transformer-based representation before the prediction layer for downstream tasks. 
Our method proposes to utilize the rule-based representations derived from a bunch of affective lexicons to improve the performance of BERT by incorporating task-specific knowledge.
The motivation for our proposal is the hypothesis that the extension of lexicon-based information will compensate for BERT's lack of proper representations of semantic and world knowledge \citep{10.1162/tacl_a_00349}, making the model more stable across domains.

\paragraph{State Space Models.}

In NLP tasks such as Part-of-Speech (POS) tagging and Named Entity Recognition, contextual information is widely acknowledged to play an important role in prediction. This leads to the adoption of structured prediction approaches such as Hidden Markov Model (HMM) \citep{rabiner1986introduction}, Maximum Entropy Markov Model (MEMM) \citep{mccallum2000maximum} and Conditional Random Field (CRF) \citep{lafferty2001CRF}, which relate a set of observable variables to a set of latent variables (e.g., words and their POS tags). 
State Space Models (SSMs) are similar to HMMs but assume continuous variables. 
Linear Gaussian SSM (LG-SSM) is a particular case of SSM in which all the conditional probability distributions are linear and Gaussian.

Following the notation from \newcite[Chap. 18]{murphy2012machine}, we briefly introduce the LG-SSM that we employ in our work.
LG-SSMs assume a sequence of observed variables $\y_{1:T}$ as input, and the goal is to draw inferences about the corresponding hidden states $\z_{1:T}$, where $T$ is the length of the sequence. Their relationship is given at each step $t$ by the equations as:
\begin{align*}\label{lgssm_trans_lyrics}
\mathbf{z}_{t}&=\mathbf {A}\mathbf {z}_{t-1} +\epsilon_t, &\epsilon_t \sim {\mathcal {N}}\left(0,\mathbf {Q}\right) \\
\mathbf{y}_{t}&=\mathbf{C}\mathbf{z}_{t} + \delta_t, &\delta_{t}\sim {\mathcal {N}}\left(0,\mathbf {R}\right)
\end{align*}
where $\Theta = (\mathbf{A}, \mathbf{C}, \mathbf{Q}, \mathbf{R})$ are the model parameters, $\epsilon_t$ is the system noise and $\delta_t$ is the observation noise. The equations above are also referred as {\em transition}\footnote{We omit control matrix $\mathbf{B}$ and control vector $\mathbf{u}_t$ in the transition equation, assuming no external influence.} and {\em observation} equations, respectively. 
Given $\Theta$ and a sequence $\y_{1:T}$, the goal is to obtain the posteriors $p(\z_t)$ for each step $t$. In an LG-SSM, this posterior is a Gaussian and can be obtained in closed form by applying the celebrated Kalman Filter \citep{kalman1960new} .



There exist some other latent variable models to estimate temporal dynamics of emotions and sentiments in product reviews \citep{mcdonald-etal-2007-structured, tackstrom2011discovering, tackstrom-mcdonald-2011-semi} and blogs \citep{DBLP:conf/aaai/KimLLP15}.
\citet{mcdonald-etal-2007-structured} and \citet{tackstrom2011discovering, tackstrom-mcdonald-2011-semi} combined document-level and sentence-level supervision as the observed variables to condition on the latent sentence-level sentiment.
\citet{DBLP:conf/aaai/KimLLP15} introduced a continuous variable $\y_t$ to solely determine the sentiment polarity $\z_t$, while $\z_t$ is conditioned on both $\y_t$ and $\z_{t-1}$ for each $t$ in the LG-SSM.

\section{Method}
\label{sec:method}

\begin{figure*}[t!]
    \centering
    \includegraphics[width=0.9\linewidth]{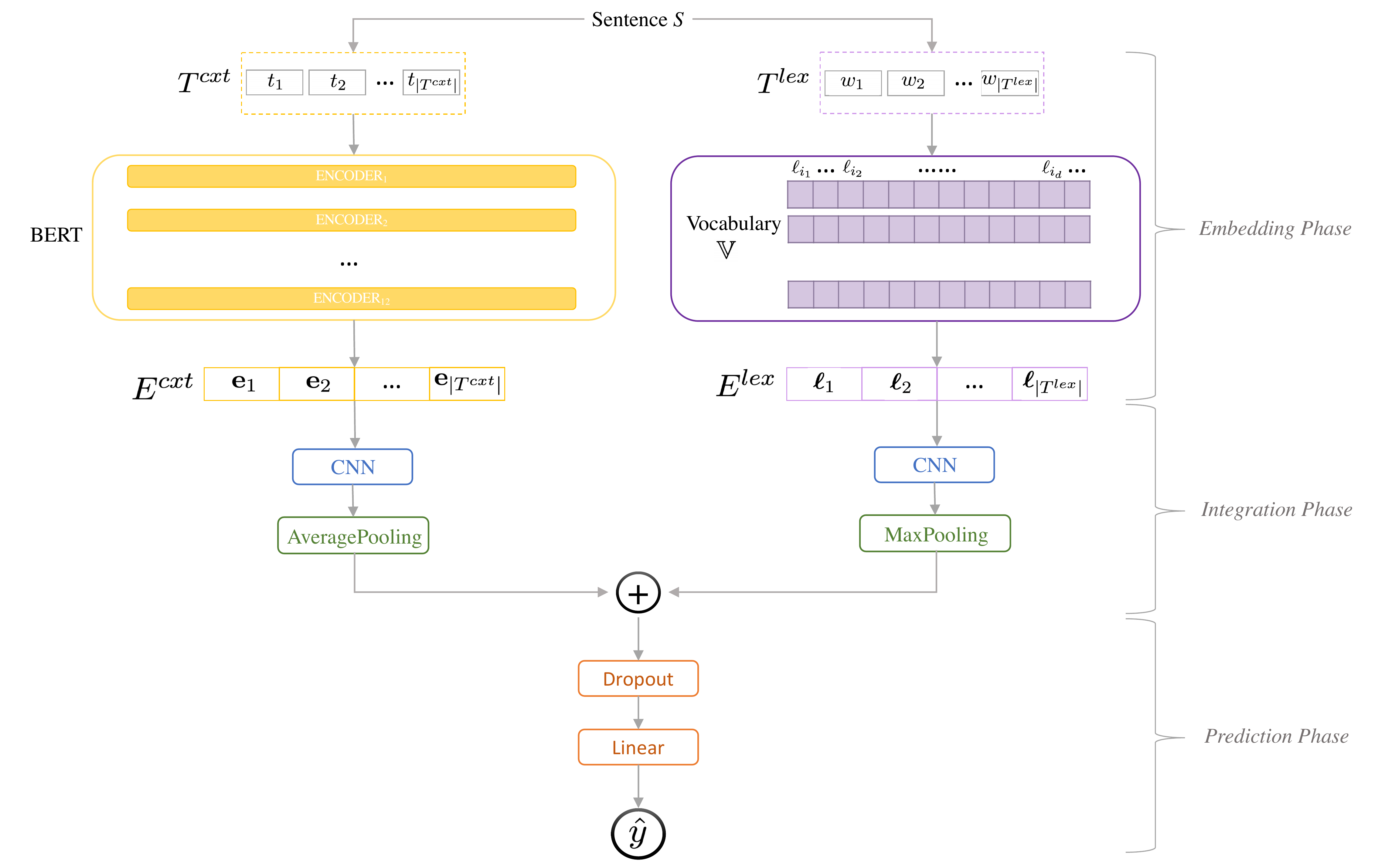}
    \caption{BERTLex architecture used for the verse-level model.}
    \label{fig:bertlex}
\end{figure*}

We propose a two-stage method to predict emotion dynamics without requiring annotated song lyrics. The first stage is a verse-level model that predicts initial scores for each song verse, where we use a hybrid approach combining lexicons and sentence-level annotated data from a different domain (\cref{sec:verse_model}). The second stage contextualizes these scores in the entire song, incorporating them into an LG-SSM trained in an unsupervised way (\cref{sec:song_model}).



\paragraph{Task Formalization.}

Let $d_x^y$ indicate the real-valued intensity of emotion $y$ for sentence/ verse $x$, where $x \in \mathcal{X}$ and $y\in \mathcal{Y}$. 
Note that $\mathcal{Y}$ = $\{y_1, y_2, \dots, y_c \}$ is a set of $c$ labels, each of which represents one of the basic emotions ($c$ = 6 for the datasets we used).
Given a source dataset $\mathbb{D}_s$ = $\{(x_1, E_1),(x_2, E_2),\dots,(x_{M}, E_{M})\}$, where $x_i$ is a sentence, $E_i$ = $\{d_{x_{i}}^{y_{1}}, d_{x_{i}}^{y_{2}}, \dots, d_{x_{i}}^{y_{c}}\}$ and $ M $ = $|\mathbb{D}_s|$.
The target dataset is $\mathbb{D}_t$ = $\{S_1, S_2, \dots, S_{|\mathbb{D}_t|} \}$, where $|\mathbb{D}_t|$ is the number of sequences (i.e., songs) and $S_i$ = $\{(v_1, E_1),(v_2, E_2),\dots,(v_{|S_i|}, E_{|S_i|})\}$ is a song consisting of $|S_i|$ verses. In the song $S_i$, the $j$-th verse $v_j$ is also associated with $c$ emotion intensities as $E_j$ = $\{d_{v_{j}}^{y_{1}}, d_{v_{j}}^{y_{2}}, \dots, d_{v_{j}}^{y_{c}}\}$. 
Given the homogeneity of label spaces of $\mathbb{D}_s$ and $\mathbb{D}_t$, the model trained by using $\mathbb{D}_s$ can be applied to predict for $\mathbb{D}_t$ directly.  
The output of verse-level model is the emotion intensity predictions $\hat{\mathbf{Y}} \in \mathbb{R}^{N \times c}$, where $N$ is the total number of verses in $\mathbb{D}_t$.
Finally, we use $\hat{\mathbf{Y}}$ as the input sequences of the song-level model to produce optimized emotion intensity sequences $\hat{\mathbf{Z}} \in \mathbb{R}^{|\mathbb{D}_t|\times c}$.


\subsection{Verse-Level Model}
\label{sec:verse_model}
Emotion lexicons provide information on associations between words and emotions \citep{ramachandran-de-melo-2020-cross}, which are proven to be beneficial in recognising textual emotions \citep{mohammad-etal-2018-semeval, DBLP:conf/coling/ZhouWWXTL20}.
Given that we would like to acquire accurate initial predictions at the verse level, we opted for a hybrid methodology that combines learning-based and lexicon-based approaches to enhance feature representation.

\paragraph{Overview.} 
The verse-level model architecture is called BERTLex, as illustrated in Figure \ref{fig:bertlex}. 
The BERTLex model consists of three phases: (1) the embedding phase, (2) the integration phase, and (3) the prediction phase. 
In the embedding phase, the input sequence is represented as both contextualized embeddings from BERT and static word embeddings from lexicons. 
In the integration phase, contextualized and static word embeddings are concatenated at the sentence level by taking the pooling operations on the two embeddings separately. 
The prediction phase encodes the integrated sequence of feature vectors and performs the verse-level emotion intensity regression by using the $\mathbb{D}_s$ as the training/development set and the $\mathbb{D}_t$ as the test set.

\paragraph{Embedding Phase.} The input sentence $S$ is tokenized in two ways: one for the pre-trained language model and the other for the lexicon-based word embedding. These two tokenized sequences are denoted as $T^{cxt}$ and $T^{lex}$, respectively. Then, $T^{cxt}$ is fed into the pre-trained BERT to produce a sequence of contextualized word embeddings $E^{cxt} = \{ \e_1, \e_2,\ldots,\e_{|T^{cxt}|}\}$, where $E^{cxt} \in \mathbb{R}^{|T^{cxt}| \times D_{cxt}}$ and $D_{cxt}$ is the embedding vector dimension. 

To capture task-specific information, a Lexicon embedding layer encodes a sequence of emotion and sentiment word associations for $T^{lex}$, generating a sequence of lexicon-based embeddings $E^{lex} = \{ \bell_1, \bell_2,\ldots,\bell_{|T^{lex}|}\}$, where $E^{lex} \in \mathbb{R}^{|T^{lex}| \times D_{lex}}$ and $D_{lex}$ is the lexical embedding vector dimension. 
We first build the vocabulary $\mathbb{V}$ from the text of $\mathbb{D}_s$ and $\mathbb{D}_t$.
For each word $\vv_i$ in $\mathbb{V}$ of $T^{lex}$, we use $d$ lexicons to generate the rule-based feature vectors $\bell_i$ = \{$\ell_{i_1}, \ell_{i_2},\ldots,\ell_{i_d}$\}, where $\ell_{i_j}$ is the lexical feature vector for word $\vv_i$ derived from the $j$-th lexicon and $D_{lex}$ = $|\bell_i|$. Additionally, we perform a degree-$p$ polynomial expansion on the feature vector $\ell_{i_j}$.

\paragraph{Integration Phase.}

As BERT uses the WordPiece tokenizer \citep{DBLP:journals/corr/WuSCLNMKCGMKSJL16} to split a number of words into a sequence of subwords, the contextualized embedding cannot be directly concatenated with the different-size static word embedding. 
Inspired by \citet{alghanmi2020combining}, we combine the contextualized embeddings and static word embeddings at the sentence level by pooling the two embeddings $E^{cxt}$ and $E^{lex}$ separately. 
To perform initial feature extraction from the raw embeddings, we apply a single-layer 1D Convolutional Neural Network \citep[CNN]{kim2014convolutional} with ReLU activation \citep{nair2010rectified} on each embedding as: 
\begin{equation*}
\small
\e_i' = \text{ReLU}(\W_1 [\e_i, \e_{i+1},\ldots,\e_{i+k-1}] + \bb_1) 
\newline
\end{equation*}
\begin{equation*}
\small
\bell_{i}'= \text{ReLU}(\W_2 [\bell_i, \bell_{i+1},\ldots,\bell_{i+k-1}] + \bb_2) 
\newline
\end{equation*}
where $\W_1$, $\bb_1$, $\W_2$ and $\bb_2$ are trainable parameters and $k$ is the kernel size. 
We then apply the average pooling and max pooling on the feature maps, respectively:
\begin{equation*}
\small
\tilde{E}^{cxt} = \text{AvgPool}(\e_1',\e_2', \ldots, \e_{|T^{cxt}|-k+1}').
\end{equation*}
\begin{equation*}
\small
\tilde{E}^{lex} = \text{MaxPool}(\bell_1',\bell_2', \ldots, \bell_{|T^{lex}|-k+1}').
\end{equation*}
Finally, the contextualised embedding and the lexicon-based embedding are merged via a concatenation layer as $ \tilde{E}^{cxt} \oplus \tilde{E}^{lex}$.

\paragraph{Prediction Phase.} 
The prediction phase outputs the emotion intensity predictions $\hat{\mathbf{Y}}$ = $\{\hat{y}_1, \hat{y}_2, \dots, \hat{y}_{N} \}$ by using a single dropout \citep{srivastava2014dropout} layer and a linear regression layer.
During training, the mean squared error loss is computed and backpropagated to update the model parameters.

\subsection{Song-Level Model}
\label{sec:song_model}

After obtaining initial verse-level predictions, the next step involves incorporating these into a song-level model using an LG-SSM.
We take one type of emotion as an example.
Specifically, we consider the predicted scores of this emotion of each song as an {\em observed} sequence $\hat{\mathbf{y}}_i$. 
That is, we group the $N$ predictions of $\hat{\mathbf{Y}}$ as $|\mathbb{D}_t|$ sequences of predictions as $\{ \hat{\mathbf{y}}_1, \hat{\mathbf{y}}_2, \dots, \hat{\mathbf{y}}_{|\mathbb{D}t|} \}$.
For the $i$-th song, the observed sequence $\hat{\mathbf{y}}_i = \y_{1:T}$ is then used in an LG-SSM to obtain the latent sequence $\hat{\z}_{1:T}$ that represents the song-level emotional dynamics, where $T$ is the number of verses in the song.

Standard applications of LG-SSM assume a temporal ordering in the sequence. This means that estimates of $p(\hat{\z}_t)$ should only depend on the observed values up to the verse step $t$ (i.e., $\y_{1:t}$), which is the central assumption to the Kalman Filter algorithm. 
Given the sequence of observations, we recursively apply the Kalman Filter to calculate the mean and variance of the hidden states, whose computation steps are displayed in Algorithm \ref{alg:kalman_filter}.

\SetKw{KwBy}{by}
\begin{algorithm}[t!]
\SetAlgoLined
\SetKwInOut{Input}{Input}
\SetKwInOut{Output}{Output}
\Input{$\mathbf{y}_t, {\hat {\mathbf {z} }}_{t-1}, \mathbf{\Sigma} _{t-1}, \mathbf {A}, \mathbf {C}, \mathbf {Q}, \mathbf {R}$ }
\Output{${\hat {\mathbf {z} }}_{t},\mathbf{\Sigma} _{t} $}
 
  \textbf{PREDICTION:} \\
  \Indp $\hat {\mathbf {z} }_{t\mid t-1} = \mathbf {A}{\hat {\mathbf {z} }}_{t-1}$\;
  $\mathbf {\Sigma} _{t\mid t-1} = \mathbf {A} \mathbf {\Sigma} _{t-1}\mathbf {A}^{\textsf {T}}+\mathbf {Q}$\;
  \Indm \textbf{MEASUREMENT:} \\
  \Indp ${\tilde {\mathbf {r} }}_{t}=\mathbf {y} _{t}-\mathbf {C}{\hat {\mathbf {z} }}_{t\mid t-1}$\;
  $\mathbf{S}_{t} = \mathbf {C}\mathbf {\Sigma} _{t\mid t-1}\mathbf {C}^{\textsf {T}}+\mathbf {R}$ \;
  $\mathbf {K} _{t}=\mathbf {\Sigma} _{t\mid t-1}\mathbf {C}^{\textsf {T}} \mathbf{S}_{t}^{-1}$ \;
  ${\hat {\mathbf {z} }}_{t}={\hat {\mathbf {z} }}_{t\mid t-1}+\mathbf {K} _{t}{\tilde {\mathbf {r} }}_{t} $\;
  $\mathbf {\Sigma} _{t} =\left(\mathbf {I} -\mathbf {K} _{t}\mathbf {C}\right)\mathbf {\Sigma} _{t|t-1}$ \;
 \Indm return ${\hat {\mathbf {z} }}_{t},\mathbf {\Sigma} _{t} $ 
 \caption{Kalman Filter}
 \label{alg:kalman_filter}
\end{algorithm}

Since we have obtained initial predictions for all verses in a song, we can assume that observed emotion scores are available for the sequence of an entire song a priori.
In other words, we can include the "future" data (i.e., $\y_{t+1:T}$) to estimate the latent posteriors $p(\hat{\z}_t)$. 
This is achieved by using the Kalman smoothing algorithm, also known as 
RTS smoother \citep{rauch1965maximum}, which is shown in Algorithm \ref{alg:kalman_smoother}. 

As opposed to most other algorithms, the Kalman Filter and Kalman Smoother algorithms are used with already known parameters.
Hence, learning the SSM involves estimating the parameters $\Theta$. If a set of gold-standard values for the complete $\z_{1:T}$ is available, they can be learned using a Maximum Likelihood Estimation (MLE). If only the noisy, observed sequences $\y_{1:T}$ are present, the Expectation-Maximization (EM) algorithm \citep{dempster1977maximum} provides an iterative method for finding the MLEs of $\Theta$ by successively maximizing the conditional expectation of the complete data likelihood until convergence.

\begin{algorithm}[t!]

\SetAlgoLined
\SetKwInOut{Input}{Input}
\SetKwInOut{Output}{Output}
\Input{$\mathbf{y}_{1:T}, \mathbf {A}, \mathbf {C}, \mathbf {Q}, \mathbf {R}$}
\Output{${\hat {\mathbf {z} }}_{t\mid T},\mathbf {\Sigma} _{t|T} $}
 \For{$t \gets 1$ \KwTo $T$ \KwBy $1$}{
 Apply the Kalman Filter (refer to Algorithm \ref{alg:kalman_filter})\;
 }
 return ${\hat {\mathbf {z} }}_{T | T},\mathbf {\Sigma} _{T | T} $ \;
 \For{$t \gets T$ \KwTo $1$ \KwBy $1$}{
 $\mathbf {J} _{t}=\mathbf {\Sigma} _{t\mid t}\mathbf {A}^{\textsf {T}}\mathbf {\Sigma} _{t+1 \mid t}^{-1}$ \;
 ${\hat {\mathbf {z} }}_{t\mid T}={\hat {\mathbf {z} }}_{t \mid t}+\mathbf {J} _{t}(\hat{\mathbf {z}} _{t+1 \mid T}-\hat{\mathbf{z}}_{t+1|t})$ \;
 $\mathbf{\Sigma} _{t|T}=\mathbf {\Sigma} _{t|t} +\mathbf {J} _{t}(\mathbf {\Sigma} _{t+1|T} - \mathbf {\Sigma} _{t+1|t}) \mathbf {J} _{t}^{\textsf {T}};$ 
 }
 return ${\hat {\mathbf {z} }}_{T : 1 \mid T},\mathbf {\Sigma} _{T:1|T}, \mathbf {J} _{T:1} $ \;
 \caption{Kalman Smoother}
 \label{alg:kalman_smoother}
\end{algorithm}

\section{Experiments}
\label{sec:experiments}


Our experiments aim to evaluate the method proposed to predict the emotional dynamics of song lyrics without utilizing any annotated lyrics data.
We introduce datasets, lexicon resources and evaluation metric used (\cref{ssec:dataset}), and discuss the implementation details and experiment settings of verse-level model (\cref{ssec:verse_exp}) and song-level model (\cref{ssec:song_exp}).


\subsection{Datasets and Evaluation}
\label{ssec:dataset}

\paragraph{LyricsEmotions.}
This corpus was developed by \citet{mihalcea2012lyrics}, consisting of 100 popular English songs with 4,975 verses in total. The number of verses for each song varies from 14 to 110. The \textsc{LyricsEmotions} dataset was constructed by extracting the parallel alignment of musical features and lyrics from MIDI tracks. These lyrics were annotated using Mechanical Turk at verse level with real-valued intensity scores ranging from 0 to 10 of six Ekman's emotions \citep{ekman1993facial}: {\fontfamily{pcr}\selectfont ANGER}, {\fontfamily{pcr}\selectfont DISGUST}, {\fontfamily{pcr}\selectfont FEAR}, {\fontfamily{pcr}\selectfont JOY}, {\fontfamily{pcr}\selectfont SADNESS} and {\fontfamily{pcr}\selectfont SURPRISE}. 
Given that our goal is to predict emotions without relying on song-level dynamics, {\em we use this dataset for evaluation purposes only.}
\paragraph{NewsHeadlines.}
To train the verse-level model, we employ the \textsc{NewsHeadlines}\footnote{\url{http://web.eecs.umich.edu/~mihalcea/affectivetext/}} dataset \citep{strapparava2007semeval}, which is a collection of 1,250 news headlines. 
Each headline is annotated with six scores ranging from 0 to 100 for each of Ekman's emotions and one score ranging from -100 to 100 for valence. 



\paragraph{Lexicons.}
\begin{table*}[t!]
\centering
\begin{tabular}{lllll}
\toprule
  & Scope     & Size (PT) & Label     & Reference                                          \\
  \midrule
\textbf{NRC-Emo-Int}   & Emotion   & 1 (4)             & Numerical & \citet{mohammad2018word}          \\
\textbf{SentiWordNet}           & Sentiment & 2 (10)            & Numerical & \citet{esuli2007sentiwordnet}     \\
\textbf{NRC-Emo-Lex}   & Emotion   & 1 (4)             & Nominal   & \citet{mohammad2013crowdsourcing} \\
\textbf{NRC-Hash-Emo}    & Emotion   & 1 (4)             & Numerical & \citet{mohammad2015using}         \\
\textbf{Sentiment140}           & Sentiment & 3 (20)            & Numerical & \citet{mohammad2013nrc}           \\
\textbf{Emo-Aff-Neg} & Sentiment & 3 (20)            & Numerical & \citet{zhu2014nrc}                \\
\textbf{Hash-Aff-Neg} & Sentiment & 3 (20)            & Numerical & \citet{mohammad2013nrc}           \\
\textbf{Hash-Senti}       & Sentiment & 3 (20)            & Numerical & \citet{kiritchenko2014sentiment}  \\
  \textbf{DepecheMood}            & Emotion   & 8 (165)           & Numerical & \citet{staiano2014depeche}        \\
  \bottomrule
\end{tabular}

\caption{Lexicons used to build lexicon-based feature vectors: PT is the feature vector size after the polynomial feature expansion.}
\label{tab:lexicons-bert}
\end{table*}

Following \citet{goel2017prayas} and \citet{meisheri-dey-2018-tcs}, we use nine emotion and sentiment related lexicons to obtain the feature vectors from the text in \textsc{NewsHeadlines} and \textsc{LyricsEmotions}, summarized in \autoref{tab:lexicons-bert}.

\paragraph{Evaluation.} 
In line with \citet{mihalcea2012lyrics}, we use the Pearson correlation coefficient ($r$) as the evaluation metric to measure the correlation between the predictions and ground truth emotion intensities.
To assess statistical significance, we conduct the Williams test \citep{williams1959comparison} in the differences between the Pearson correlations of each pair of models.

For baselines, our method is unsupervised at the song level, and we are not aware of prior work tackling similar cases. Therefore, we use the results of the verse-level model as our main baseline. We argue that this is a fair baseline since the SSM-based model does not require additional data.

\subsection{Verse-level Experiments}
\label{ssec:verse_exp}

\paragraph{Setup.}
For the pre-trained model, we choose the BERT$_{base}$ uncased model in English with all parameters frozen during training.
All models are trained on an NVIDIA T4 Tensor Core GPU with CUDA (version 11.2).

\paragraph{BERTLex.}

The sequence of embeddings for each token, including [CLS] and [SEP] at the output of the last layer of the BERT$_{base}$ model, is fed into a Conv1D layer with 128 filters and a kernel size of 3, followed by a 1D global average pooling layer.

We concatenate nine vector representations for every word in the established vocabulary by using the lexicons in \autoref{tab:lexicons-bert} in the identical order to form a united feature vector. As a result, the whole word embedding is in the shape of (3309,25), where 3309 is the vocabulary size and 25 is the number of lexicon-based features. To validate if adding polynomial features can make better predictions, we also perform a polynomial feature expansion with a degree of 3, extending the shape of vector representations to (3309, 267). Then, static word embeddings are fed a Conv1D layer with 128 filters and a kernel size of 3, followed by a global max-pooling layer. 

The two pooled vectors are then concatenated through a Concatenate layer as they are in the same dimensionality.
We generate the predictions of emotion intensities by using a Linear layer with a single neuron\footnote{We experimented with a multi-task model that predicted all six emotions jointly, but preliminary results showed that building separate models for each emotion performed better.} for regression.



\begin{table}[t!]
\centering
\begin{tabular}{@{}l|l@{}}
\toprule
Parameters    & Value \\ \midrule
Dropout rate  & 0.1   \\ 
Optimizer     & Adam  \\ 
Learning rate & 2e-5  \\ 
$\beta_1$     & 0.9   \\ 
$\beta_2$     & 0.999 \\ 
Batch size    & 32    \\

\bottomrule
\end{tabular}
\caption{Hyperparameter settings of BERT and CNN models.}
\label{tab:hyper-para}
\end{table}

\paragraph{Training.}

Instead of using the standard train/dev/test split of the \textsc{NewsHeadlines} dataset, we apply 10-fold cross-validation to tune the hyper-parameters of BERT-based models. Empirically tuned hyperparameters are listed in Table \ref{tab:hyper-para} and are adopted in the subsequent experiments unless otherwise specified. After tuning, the final models using this set of hyperparameters are trained on the full \textsc{NewsHeadlines} data. We use an ensemble of five runs, taking the mean of the predictions as the final output.


\begin{table*}[t!]
\centering
\begin{tabular}{@{}llllllll@{}}
\toprule
  \toprule  & Dataset 
  & {\fontfamily{pcr}\selectfont ANG}  & {\fontfamily{pcr}\selectfont DIS} & {\fontfamily{pcr}\selectfont FEA}   &{\fontfamily{pcr}\selectfont JOY}    & {\fontfamily{pcr}\selectfont SAD} & {\fontfamily{pcr}\selectfont SUR}   \\ \midrule \midrule
Lexicon only    & \sc{NH}  & 0.197 & 0.106 & 0.231 & 0.219  & 0.112 & 0.056 \\
                & \sc{LE} & 0.212 & 0.091 & 0.185 & 0.209  & 0.175 & 0.031 \\ \midrule
BERT only       & \sc{NH}  & 0.740 & 0.651 & 0.792 & 0.719 & 0.808 & 0.469 \\
                & \sc{LE} & 0.311 & 0.261 & 0.314 & 0.492  & 0.306 & 0.071 \\ \midrule \midrule
BERTLex         & \sc{NH}  & 0.865 & 0.828 & 0.840 & 0.858  & 0.906 & 0.771 \\
                & \sc{LE} & 0.340 & \textbf{0.287} & 0.336 & 0.472  & 0.338 & 0.066 \\ \midrule
BERTLex$^{poly}$ & \sc{NH}  & 0.838 & 0.788 & 0.833 & 0.840  & 0.885 & 0.742 \\
                & \sc{LE} & \textbf{0.345} & 0.268 & \textbf{0.350} & \textbf{0.503}  & \textbf{0.350} & \textbf{0.089} \\ \bottomrule
\end{tabular}
\caption[The LOF caption]{Pearson correlations between gold-standard labels and predictions of the verse-level models in the \textsc{NewsHeadlines (NH)} and \textsc{LyricsEmotions (LE)} datasets.}
\label{tab:bert_results}
\end{table*}

\subsection{Song-Level Experiments}
\label{ssec:song_exp}

We apply the library  {\fontfamily{pcr}\selectfont pykalman} (version 0.9.2)\footnote{\url{https://github.com/pykalman/pykalman}}, which implements the Kalman Filter, the Kalman Smoother and the EM algorithm to train SSMs. We fix the initial state mean as the first observed value in the sequence (i.e., each song's first verse-level prediction) and the initial state covariance as 2. 
We then conduct experiments with several groups of parameters transition matrices $\mathbf{A}$, transition covariance $\mathbf{Q}$, observation matrices $\mathbf{C}$ and observation covariance $\mathbf{R}$ to initialise the Kalman Filter and Kalman Smoother. 
For parameter optimization, we experiment n\_iter = \{1,3,5,7,10\} to control the number of EM algorithm iterations. 
Additionally, we apply 10-fold cross-validation when choosing the optimal parameters via EM, which means each song is processed by a Kalman Filter or Kalman Smoother defined by the optimal parameters that we obtained from training on the other 90 songs.

\section{Results and Analysis}
In this section, we report and discuss the results of the experiments. We first compare the results of our lexicon-based, learning-based and hybrid methods at the verse level (\cref{sec:results_verse}).
We then provide the results of song-level models and investigate the impact of initial predictions from verse-level models, SSM parameters, and parameter optimization (\cref{sec:results_ssm}). 
We additionally show the qualitative case analysis results to understand our model's abilities and shortcomings (\cref{sec:case_study}).

\subsection{Results of Verse-level Models}
\label{sec:results_verse}

\autoref{tab:bert_results} shows the results of verse-level models on the \textsc{NewsHeadlines} (average of 10-fold cross-validation) and \textsc{LyricsEmotions} (as the test set) datasets. 
The domain difference is significant in news and lyrics, as we can observe from the different performances of BERT-based models on the two datasets.
Overall, our BERTLex method outperforms the lexicon-only and BERT-only baselines and exhibits the highest Pearson correlation of 0.503 (BERTLex$^{poly}$ for {\fontfamily{pcr}\selectfont JOY}) in \textsc{LyricsEmotions}.

Having a closer look at the results of \textsc{LyricsEmotions}, we also observe the following:
\begin{itemize}
\item The addition of lexicons for incorporating external knowledge consistently promotes the performance of BERT-based models.
\item BERTLex models with polynomial feature expansion are better than those without, except for {\fontfamily{pcr}\selectfont DISGUST}.
\item Our models are worst at predicting the emotion intensities of {\fontfamily{pcr}\selectfont SURPRISE} (lower than 0.1), which is in line with similar work in other datasets annotated with the Ekman taxonomy.
\end{itemize}

\subsection{Results of Song-level Models}
\label{sec:results_ssm}

Extensive experiments confirm that our song-level models utilizing the Kalman Filter and Kalman Smoother can improve the initial predictions from verse-level models combining BERT and lexicons (see \autoref{tab:SSM_results_poly} and \autoref{tab:SSM_results_params_trans}). The LG-SSMs with EM-optimized parameters always perform better than those without using EM. 
Furthermore, the performance improvements of the strongest SSMs from their corresponding verse-level baselines are statistically significant at 0.05 confidence (marked with *), except for {\fontfamily{pcr}\selectfont SURPRISE}. 

Theoretically, the Kalman Smoother is supposed to perform better than the Kalman Filter since the former utilizes all observations in the whole sequence. According to our experimental results, however, the best-performing algorithm depends on emotion. On the other hand, running the EM algorithm consistently improves the results of SSMs that simply use the initial values, except for {\fontfamily{pcr}\selectfont SURPRISE}.

\begin{table*}[t!]
\centering
\npdecimalsign{.}
\nprounddigits{3}
\begin{tabular}{lN{0}{3}N{0}{3}N{0}{3}N{0}{3}N{0}{3}N{0}{3}}
  \toprule
  & {\fontfamily{pcr}\selectfont ANG}  & {\fontfamily{pcr}\selectfont DIS} & {\fontfamily{pcr}\selectfont FEA}   &{\fontfamily{pcr}\selectfont JOY}    & {\fontfamily{pcr}\selectfont SAD} & {\fontfamily{pcr}\selectfont SUR} \\
  \midrule
  \midrule

  BERTLex & 0.3383 & 0.2796  & 0.3356 & 0.4678 & 0.3377  & 0.0656   \\
  \midrule
Filter & 0.3594$^*$ & 0.2865$^*$  & 0.3523$^*$ & 0.4982$^*$ & 0.3612$^*$  & {\npbold} 0.0691   \\
Smoother & 0.3615$^*$ & 0.2822  & 0.3523$^*$ & 0.5006$^*$ & 0.3657$^*$  & 0.0635   \\
Filter-EM & 0.3960$^*$ & {\npbold} 0.2928$^*$  & {\npbold} 0.3568$^*$ & 0.5222$^*$ & {\npbold} 0.3871$^*$  & 0.0688   \\
  Smoother-EM  & {\npbold} 0.4050$^*$ & 0.2797  & 0.3392 &{\npbold} 0.5224 $^*$ & 0.3849$^*$  & 0.0596   \\
  \midrule
  \midrule

  BERTLex$^{poly}$    & 0.3146 & 0.2613  & 0.3501 & 0.5034 & 0.3466  & 0.0831   \\ 

\midrule
Filter   & 0.3343$^*$ & 0.2667  & 0.3672$^*$ & 0.5384$^*$ & 0.3740$^*$  & {\npbold} 0.0876   \\
Smoother   & 0.3315$^*$ & 0.2579  & 0.3677$^*$ & 0.5422$^*$ & 0.3804$^*$  & 0.0823   \\
Filter-EM     & {\npbold} 0.3581$^*$ & {\npbold}0.2696$^*$  & {\npbold}0.3713$^*$ & 0.5679$^*$ & 0.4051$^*$  & 0.0870   \\
Smoother-EM   & 0.3564$^*$ & 0.2506  & 0.3551 & {\npbold}0.5697$^*$ & {\npbold}0.4052$^*$  & 0.0786   \\ \bottomrule
\end{tabular}
\caption{Pearson correlations between gold-standard emotion intensities and predictions of BERTLex models and SSMs, respectively. The default parameter settings in  {\fontfamily{pcr}\selectfont pykalman} are used: $\mathbf{A}$ = 1, $\mathbf{Q}$ = 1, $\mathbf{C}$ = 1, $\mathbf{R}$ = 5 and n\_iter = 10. }
\label{tab:SSM_results_poly}
\end{table*}

\begin{table*}[t!]
\centering
\npdecimalsign{.}
\nprounddigits{3}
\begin{tabular}{lN{0}{3}N{0}{3}N{0}{3}N{0}{3}N{0}{3}N{0}{3}}
  
\toprule
\multicolumn{1}{l}{}                                   
  & {\fontfamily{pcr}\selectfont ANG}  & {\fontfamily{pcr}\selectfont DIS} & {\fontfamily{pcr}\selectfont FEA}   &{\fontfamily{pcr}\selectfont JOY}    & {\fontfamily{pcr}\selectfont SAD} & {\fontfamily{pcr}\selectfont SUR} \\
  \midrule
  \midrule

\multicolumn{7}{l}{$\mathbf{A}$ = 0.5} \\
\multicolumn{1}{l}{Filter}                                    & 0.2645 & 0.2231  & 0.3504 & 0.4545 & 0.3507  & 0.0736   \\
\multicolumn{1}{l}{Smoother}                                  & 0.2767 & 0.2225  & 0.3512 & 0.4709 & 0.3624$^*$  & 0.0707   \\
\multicolumn{1}{l}{Filter-EM} & 0.3565$^*$ & {\npbold}0.2726$^*$  & {\npbold}0.3725$^*$ & 0.5627$^*$ & 0.3973$^*$  & {\npbold}0.0885   \\
  \multicolumn{1}{l}{Smoother-EM}                               & {\npbold}0.3600$^*$ & 0.2619  & 0.3635$^*$ & {\npbold}0.5687$^*$ & {\npbold}0.4022$^*$  & 0.0825   \\
  \midrule
\multicolumn{7}{l}{$\mathbf{A}$ = 2}                                    

\\
\multicolumn{1}{l}{Filter}                                    & 0.3542$^*$ & 0.2701  & 0.3693$^*$ & 0.5595$^*$ & 0.3931$^*$  & 0.0849   \\
\multicolumn{1}{l}{Smoother}                                  & 0.0745 & 0.0545  & 0.1603 & 0.2045 & 0.1739  & 0.0034   \\
\multicolumn{1}{l}{Filter-EM}             & 0.3547$^*$ & {\npbold}0.2718$^*$  & {\npbold}0.3746$^*$ & 0.5623$^*$ & 0.3988$^*$  & {\npbold}0.0888   \\
\multicolumn{1}{l}{Smoother-EM}                               & {\npbold}0.3582$^*$ & 0.2604  & 0.3641$^*$ & {\npbold}0.5682$^*$ & {\npbold}0.4031$^*$  & 0.0827   \\ \bottomrule
\end{tabular}

\caption{Pearson correlations between gold-standard labels and SSMs with different values of transition matrices $\mathbf{A}$, on the basis of BERTLex$^{poly}$ models (as listed in the bottom half of \autoref{tab:SSM_results_poly}). The other parameters are fixed as $\mathbf{Q}$ = 1, $\mathbf{C}$ = 1, $\mathbf{R}$ = 5 and n\_iter = 5. }
\label{tab:SSM_results_params_trans}
\end{table*}

\paragraph{Impact of verse-level predictions.}  
The performances of applying Kalman Filter, Kalman Smoother and EM algorithm are associated with the initial scores predicted by verse-level models. 
For the same emotion, we compare the results based on the mean predictions from the BERTLex models with and without polynomial expansion on lexical features, respectively (shown in \autoref{tab:SSM_results_poly}). 
We observe that the higher the Pearson correlation between the ground truth and the verse-level predictions, the more accurate the estimates obtained after using LG-SSMs accordingly. 
The strongest SSMs also differ with the different emotion types and initial predictions, as denoted in boldface. 

\paragraph{Impact of initial parameters.} 
The results of Kalman Filter and Kalman Smoother are sensitive to the initial model parameters. As displayed in \autoref{tab:SSM_results_params_trans}, when we only change the value of transition matrices $\mathbf{A}$ and fix the other parameters, the performance of Kalman Filter and Kalman Smoother can be decreased even worse than without them. Fortunately, this kind of diminished performance due to the initial parameter values can be diluted by optimizing the parameters with an EM algorithm.

\paragraph{Impact of parameter optimization.} 
For either Kalman Filter or Kalman Smoother, using the EM algorithm to optimize the parameters increases Pearson's $r$ in most cases.
Through experiments, the number of iterations does not significantly influence the performance of the EM algorithm, and 5 $\sim$ 10 iterations usually produce the strongest results.

\subsection{Qualitative Case Studies}
\label{sec:case_study} 

To obtain some insights into further improvement, we examine the errors that our models are making.

\begin{figure}[t!]
  \centering
  \includegraphics[width=\linewidth]{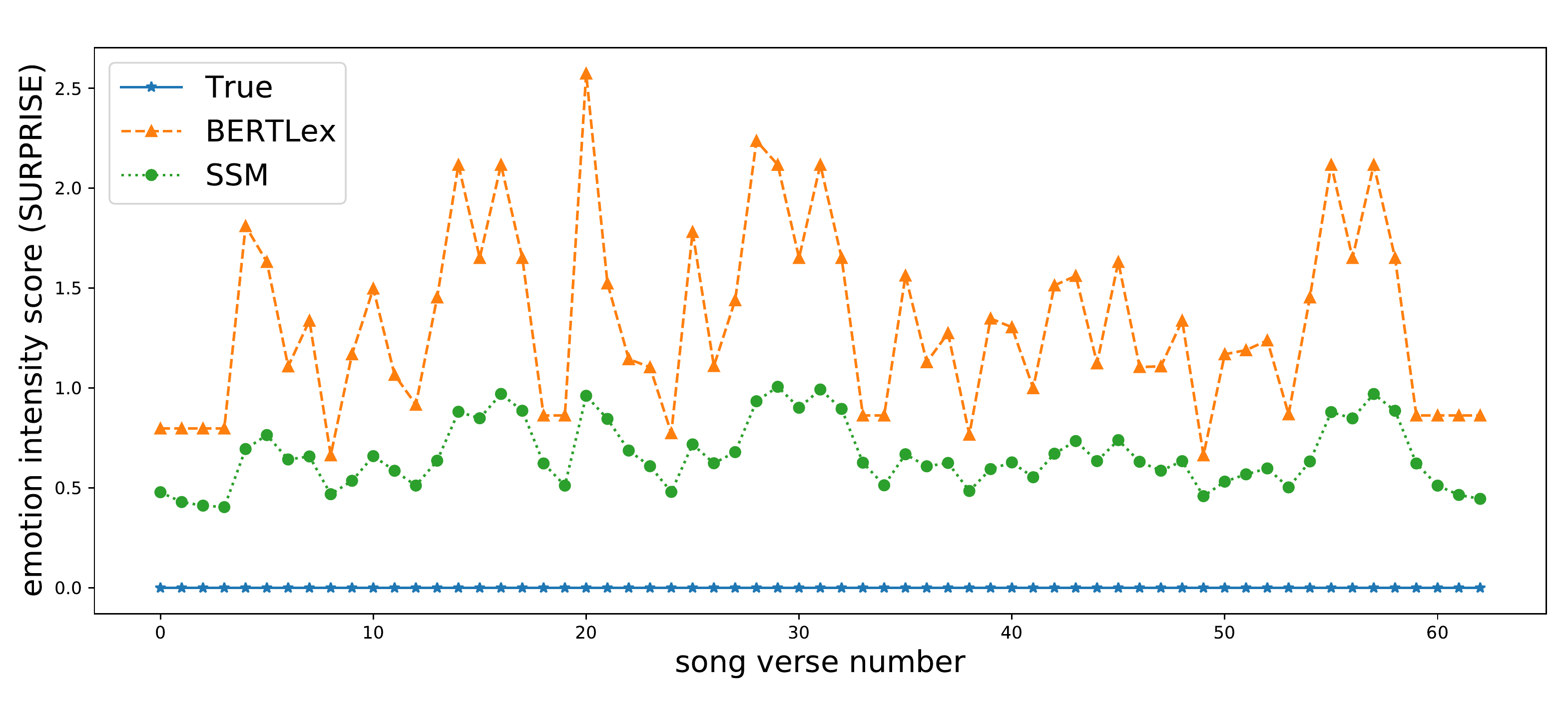}
  \caption{The {\fontfamily{pcr}\selectfont SURPRISE} emotion intensities of ground truth (all zeros), BERTLex model and SSM in an example song.}
  \label{fig:surprise_zeros_image}
\end{figure}

\paragraph{Domain discrepancy.} 
As displayed in Section \ref{sec:results_ssm}, the Pearson correlations between the ground-truth labels and estimates of {\fontfamily{pcr}\selectfont SURPRISE} are lower than 0.1, which means our verse-level and song-level models both underperform in predicting the intensities of this emotion. With closer inspection, we observe that there are a great number of zeros (annotated as 0) in the ground-truth annotations of {\fontfamily{pcr}\selectfont SURPRISE} in the target domain dataset. For example, \autoref{fig:surprise_zeros_image} shows the emotion dynamic curves of \textit{If You Love Somebody Set Them Free} by Sting, where all gold-standard labels of {\fontfamily{pcr}\selectfont SURPRISE} are zeros in the whole song. Statistically, there are 1,933 zeros out of 4,975 (38.85\%) {\fontfamily{pcr}\selectfont SURPRISE} gold-standard labels in \textsc{ LyricsEmotions} but only 148 of 1,250 (11.84\%) zeros in \textsc{NewsHeadlines}. 
The models trained on \textsc{NewsHeadlines} would not assume such a large absences of {\fontfamily{pcr}\selectfont SURPRISE} when predicting for \textsc{ LyricsEmotions}. This apparent domain discrepancy clearly affects the performance of our method.

\begin{figure}[t!]
   \centering
   \includegraphics[width=\linewidth]{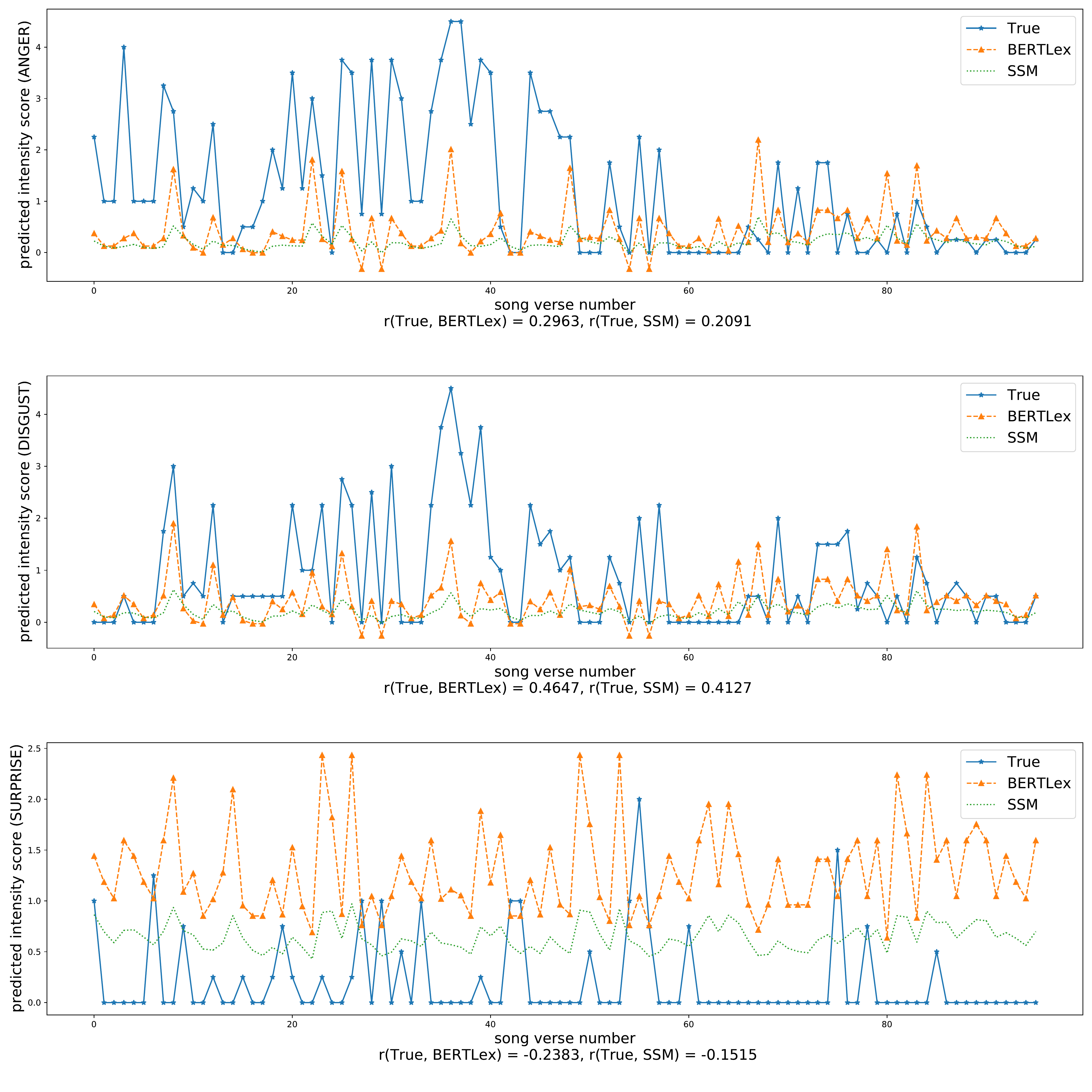}
   \caption{Emotional dynamics of {\fontfamily{pcr}\selectfont ANGER}, {\fontfamily{pcr}\selectfont DISGUST} and {\fontfamily{pcr}\selectfont SURPRISE} in \textit{Bad Romance} by Lady Gaga: Pearson's $r$ between ground truth and predictions of BERTLex$^{poly}$, estimates of Kalman Filter, are reported, respectively.}
   \label{fig:fluctuate_image}
\end{figure}

\paragraph{Characteristics of Kalman Filter and Kalman Smoother.}
Our experiments indicate that initial predictions of at least 50 to 70 of 100 songs have been enhanced after modelling them with LG-SSMs. We summarize two trending types from the emotional dynamics of the songs whose predictions are weakened by LG-SSMs. 
One is that the gold-standard emotional dynamics fluctuate more sharply than those of the verse-level predictions, as displayed in the first and the second sub-figures in \autoref{fig:fluctuate_image}. The other is the opposite that verse-level models produce an emotion intensity curve with more sudden changes than the gold-standard emotional dynamics (see the third sub-figure in \autoref{fig:fluctuate_image}). 
The emotional dynamics trend of estimates by song-level models is similar to verse-level models. 
Due to the Gaussian assumption, Kalman Filter and Kalman Smoother tend to flatten or smooth the curves of verse-level predictions.
This means that applying LG-SSMs can somewhat reduce errors in the second type of emotion dynamic curves. For the first type, however, the Kalman Filter and Kalman Smoother make the results worse, as smoother estimations are not desirable in this situation.

\paragraph{Using text solely.} 

Lyrics in \textsc{LyricsEmotions} are synchronised to acoustic features, where some verses with identical text are labelled as different emotional intensities. For instance, in \autoref{tab:error_same},  the verse \textit{"When it rain and rain, it rain and rain"} repeats multiple times in the song \textit{Rain} by Mika, and their gold-standard {\fontfamily{pcr}\selectfont SADNESS} labels differ in different verses as the emotion progresses with music. 
However, the verse-level models can only produce the same predictions since these verses share exactly the same text, and the models do not consider the context of the whole song.
Consequently, the emotion scores of different verses predicted by LG-SSMs are close, as the results of song-level models are highly related to the initial predictions from BERTLex.

\begin{table}[t!]
\centering
\small
\begin{tabular}{llll}
\toprule
  Verse ID & Gold & BERTLex & Smoother-EM \\
  \midrule
s55v15  & 4.33  & 8.65    & 1.68    \\
s55v31  & 7.66  & 8.65    & 1.68    \\
s55v32  & 7.33  & 8.65    & 1.63    \\ 
\bottomrule
\end{tabular}
\caption{{\fontfamily{pcr}\selectfont SADNESS} scores of verses with the same lyrics verse \textit{ "When it rain and rain, it rain and rain" } but different gold-standard labels in the song.}
\label{tab:error_same}
\end{table}

\section{Conclusion and Future Work}

This paper presents a two-stage BERTLex-SSM framework for the sequence-labelling emotion intensity recognition tasks. 
Combining the contextualized embeddings with static word embeddings and then modelling the initial predicted intensity scores as a State Space model, our method can utilize context-sensitive features with external knowledge and capture the emotional dynamic transitions.
Experimental results show that our proposed BERTLex-SSM effectively predicts emotion intensities in the lyrics without requiring annotated lyrics data.

Our analysis in Section \ref{sec:case_study} points to a range of directions for future work:
\begin{itemize}
  \item  While our method could apply any general verse-level model, including a pure lexicon-based one, in practice, we obtained the best results by leveraging annotated sentence-level datasets. This naturally leads to the domain discrepancy: in our particular case, between the news and lyrics domains. Given that unlabelled song lyrics are relatively easy to obtain, one direction is to incorporate unsupervised domain adaptation techniques \citep{ramponi-plank-2020-neural} to improve the performance of the verse-level model. Semi-supervised learning (similar to \citet{tackstrom-mcdonald-2011-semi}) is another promising direction in this avenue, although methods would need to be modified to incorporate the continuous nature of the emotion labels.
  \item Despite being able to optimize the estimates through Kalman Filter and Kalman Smoother, the simplicity of the LG-SSM makes it difficult to deal with the wide variations in emotion space dynamics, given that it is a linear model. We hypothesize that non-linear SSM extensions \citep{julier1997new, ito2000gaussian, julier2004unscented} might be a better fit for modelling emotion dynamics.
  \item As the \textsc{LyricsEmotions} dataset is annotated on parallel acoustic and text features, using lyrics solely as the feature can cause inconsistencies in the model. Extending our method to a multi-modal setting would remedy this issue when the identical lyrics are companions with different musical features to appear in various verses. Taking the knowledge of song structure (e.g., Intro - Verse - Bridge - Chorus) into account has the potential to advance the recognition of emotion dynamics, assuming the way (up or down) that emotion intensities change is correlated with which part of the song the verses locate.

\end{itemize}

\bibliography{tacl2021}
\bibliographystyle{acl_natbib}

\end{document}